\documentclass[11pt,a4paper]{article}
\usepackage[margin=0.83in,headheight=14pt]{geometry}
\usepackage[T1]{fontenc}
\usepackage[utf8]{inputenc}
\usepackage{tgpagella}
\usepackage[scale=.95]{tgheros}
\usepackage{inconsolata}
\usepackage{graphicx}
\usepackage{booktabs,longtable,tabularx,array,calc}
\usepackage{float}
\usepackage[font=small,labelfont=bf,justification=raggedright,singlelinecheck=false]{caption}
\usepackage{amsmath,amssymb}
\usepackage{microtype}
\usepackage{enumitem}
\setlist{nosep,topsep=4pt}
\usepackage{xcolor}
\definecolor{HGCNavy}{HTML}{20374E}
\definecolor{HGCTeal}{HTML}{287977}
\usepackage{tikz}
\usetikzlibrary{arrows.meta,positioning,calc,fit,backgrounds}
\definecolor{HGCTealLight}{HTML}{E8F2F1}
\definecolor{HGCPanel}{HTML}{F4F6F8}
\definecolor{HGCLine}{HTML}{B9C1C9}
\definecolor{HGCMuted}{HTML}{5F6B77}
\newcommand{\hgcFont}{\fontsize{8}{9.6}\selectfont}
\newcommand{\hgcSmall}{\fontsize{7}{8.4}\selectfont}
\newcommand{\hgcMark}{\fontsize{9}{10}\selectfont\bfseries}
\tikzset{
  hgc/.style={font=\hgcFont, text=HGCNavy, line cap=round, line join=round},
  box/.style={draw=HGCLine, fill=HGCPanel, rounded corners=2.5pt, line width=0.6pt,
              align=left, inner xsep=6pt, inner ysep=4.5pt, anchor=center},
  key/.style={box, draw=HGCTeal, fill=HGCTealLight, line width=0.9pt},
  ghost/.style={box, draw=HGCLine, fill=white, dashed, text=HGCMuted},
  flow/.style={-{Stealth[length=5pt,width=4pt]}, line width=0.75pt, draw=HGCNavy},
  tflow/.style={flow, draw=HGCTeal, line width=0.9pt},
  mflow/.style={flow, draw=HGCMuted, dashed},
  lbl/.style={font=\hgcSmall, text=HGCMuted, inner sep=1.5pt, align=center},
  ttl/.style={font=\hgcFont\bfseries, text=HGCNavy},
}
\newcommand{\hgcHead}[1]{\textbf{#1}}
\newcommand{\hgcNote}[1]{{\hgcSmall\color{HGCMuted}#1}}

\newcommand{\defn}[1]{\textit{#1}}
\newcolumntype{H}{>{\raggedright\arraybackslash}X}
\newcolumntype{N}[1]{>{\raggedright\arraybackslash}p{#1}}
\renewcommand{\arraystretch}{1.13}
\usepackage{xurl}
\usepackage{hyperref}
\hypersetup{colorlinks=true,linkcolor=HGCNavy,urlcolor=HGCNavy,citecolor=HGCNavy}
\usepackage{titlesec}
\titlespacing*{\section}      {0pt}{2.6ex plus 0.8ex minus 0.2ex}{1.0ex plus 0.2ex}
\titlespacing*{\subsection}   {0pt}{2.2ex plus 0.6ex minus 0.2ex}{0.8ex plus 0.2ex}
\titlespacing*{\subsubsection}{0pt}{1.8ex plus 0.5ex minus 0.2ex}{0.6ex plus 0.1ex}
\usepackage{fancyhdr}
\title{HGC: Preserving Organizational Control When Switching AI Agent Providers}
\author{Amjad Yousef Majid\\\href{mailto:amjad@slimx.ai}{amjad@slimx.ai}}
\date{November 2026}
\begin{document}
\maketitle
\begin{abstract}
Organizations increasingly use AI agents from different providers and need the freedom to replace them. Yet connecting a new agent to the same tools does not ensure that it follows the same organizational rules. One system may treat an approval as permission for a single action, while another may extend it to an entire session. A failed call may also have changed an external system even when no result was recorded. We propose the Host Governance Contract (HGC), a provider-independent agreement that keeps organizational authority and execution history under the control of a host—the system responsible for enforcing organizational policies and maintaining execution records—while allowing the agent runtime to change. HGC defines five linked records and eight obligations covering authorization of individual actions, controlled execution, uncertain outcomes, and provider handover. We illustrate the contract through a procurement scenario, discuss its application to software delivery and protected-data disclosure, compare existing agent tools and enterprise platforms, present conditional safety arguments, and propose 13 conformance tests. HGC relies on trusted enforcement, complete mediation of governed actions, and crash-safe execution state. It does not guarantee exactly-once effects in external services. Its central contribution is a precise criterion for replacing AI agent runtimes while preserving organizational control and the meaning of prior decisions, execution records, and unresolved actions.

\end{abstract}
\noindent\textbf{Keywords:} AI agents; agent governance; provider portability; authorization; interoperability.

\section{Introduction}\label{introduction}

Organizations increasingly depend on AI agents to write code, examine business records, and act in external systems. They may use one provider today and another next month, or several providers at once. Switching is attractive when a different agent is more capable, less expensive, or better suited to the organization's data requirements. The technical connection is becoming easier. The Model Context Protocol (MCP) connects agents with tools and resources; Agent2Agent (A2A) supports communication between agents; and the Agent Client Protocol connects coding agents with development environments \cite{ref1, ref2, ref3}. But using the same interface does not mean following the same organizational rules. 
Consider three ordinary kinds of work. A procurement agent prepares an order that, under the organization's policy, needs a manager's approval. A coding agent can inspect a repository but may not deploy to production without authorization. A research agent reads confidential documents that must not be sent to an unapproved model provider. A replacement agent might support all the same tools and still create a problem: it might interpret approval more broadly, repeat an order after a timeout, or disclose documents to a newly selected service.
The procurement case makes the problem concrete. Suppose a manager approves an order for a specified supplier, amount, and delivery address. If the agent changes the amount or recipient, the original approval should not automatically follow it. If the ordering service receives the order but the response is lost, the order may already exist, and sending it again could create a duplicate. If another agent then takes over, it has no reliable way of knowing that the first order may have gone through. The same questions arise with software deployments and other external actions, though their effects and recovery methods differ. 
Existing products already offer many of the necessary controls. Agent tools implement permission prompts and sandboxes, while enterprise platforms manage identity, tool policies, and logs \cite{ref4, ref5, ref6, ref7, ref8, ref9, ref10}. What is missing is an explicit agreement about what these controls mean and what survives when the runtime changes. An ``approved'' record should not quietly change from one exact order to a whole class of operations, and an unresolved external effect should not disappear because a new model starts a session. 

\noindent\textbf{The separation we propose.}\ We distinguish the \defn{runtime}, which plans and coordinates an agent's work, from the \defn{host}, which controls organizational authority, access to protected resources, and the authoritative execution record. A runtime may be an agent harness supplied by a provider. A host may be an enterprise platform, gateway, editor, or control room. One product can implement both roles. HGC is an agreement between the roles. The central rule is simple:

\begin{quote}
\emph{The runtime proposes what to do; the host decides what may happen and records what is known to have happened.}
\end{quote}

We call this agreement the \defn{Host Governance Contract} (HGC). It does not prescribe a model, planner, API, or database. For the three scenarios above, it has five practical consequences:

\begin{enumerate}
\item \textbf{Tool access is not authority.} Being able to call the deployment or ordering tool does not itself permit the action.
\item \textbf{Standing permission and action approval serve different purposes.} An employee may submit routine orders within an authorized spending limit. The organization may require separate manager approval for higher-value orders; HGC preserves whichever rule the host applies.
\item \textbf{A decision covers only the action it was made for.} A material change to the order, deployment, or data  requires a new decision, whether that decision comes from a person or from policy.
\item \textbf{Authorization and execution are separate events.} Immediately before dispatch, the host checks current authority again and permits at most one host-side send of that identified action.
\item \textbf{An absent result is not a failed action.} The host retains the uncertainty and restricts potentially conflicting work across crashes and handover, subject to explicit, recorded exceptions.
\end{enumerate}

The formal records and obligations appear in Sections 4 and 5. A host can guarantee these restrictions only for effects that actually pass through its trusted enforcement path; it cannot govern a private credential or network route that bypasses that path.

\noindent\textbf{What HGC contributes.}\ First, Section 2 compares documented approval, policy, and recovery controls to show why their different meanings matter to integration. Second, Sections 3--5 define five host-linked records and eight obligations, labeled H1--H8 (H for host), separating enforcement the host must provide from promises made by a cooperative runtime. Third, Section 6 states six conditions for governance-preserving provider replacement, gives three conditional safety arguments, and defines tests that could check independent implementations.

The individual mechanisms have prior art, including action-decision protocols and durable approval designs (Section 7). HGC's contribution is to bring them together around one requirement: preserving organizational control when the agent runtime changes. The host keeps its authority, decisions keep their meaning, and unresolved execution state survives the change of provider. This paper specifies the design and argues for it; it does not yet report an implementation or measurements, and Section 9 discusses what that leaves open.

\section{Existing Controls and Why Their Meanings Matter}\label{existing-controls-and-why-their-meanings-matter}

We reviewed public documentation for three agent execution tools---DeepSeek Harness, Claude Code, and OpenAI Codex---and four enterprise platforms: Microsoft Agent 365, Amazon Bedrock AgentCore, OpenAI Workspace Agents, and Salesforce Agentforce. We included Microsoft's Agent Governance Toolkit approval design as a separate design example. We do not infer that a system lacks a control merely because its public documentation does not describe it. \\

\noindent\textbf{Agent tools control individual operations.} DeepSeek Harness describes a one-time grant for a tool call \cite{ref4}. Claude Code combines configurable permission modes with a sandbox that limits what commands may reach \cite{ref5, ref11}. Codex combines an approval policy and sandbox; its documented workflows include individual approvals and broader session-level permission, while some review can come from an automated reviewer \cite{ref6, ref12}. These controls differ in scope, in the source of the decision, and in the component that checks it. A host must preserve those distinctions when its runtime changes. \\

\noindent\textbf{Enterprise platforms manage organizational policy and visibility.} Enterprise agent platforms provide controls for identity, tool access, authorization, and oversight. Microsoft Agent 365 documents agent identity, lifecycle management, and centralized management of available tools \cite{ref7, ref13, ref14}. Amazon Bedrock AgentCore evaluates tool invocations against gateway policies and distinguishes enforcement from log-only evaluation \cite{ref8, ref15, ref16}. OpenAI Workspace Agents describes role-based access, approval gates, and logs \cite{ref9}, while Salesforce Agentforce documents lifecycle practices and session trace export \cite{ref10, ref17}. These controls can support a host's governance of agent work. HGC specifies what must carry across a runtime change: the link between the exact operation, the authority under which it was allowed, and the evidence of what happened. The host can then keep enforcing those decisions and unresolved restrictions when a replacement runtime takes over. \\

\noindent\textbf{The critical distinction is scope.} Table 1 asks the same questions of five documented designs: what does a favorable decision cover, and where is that decision checked? Human approval and machine policy are deliberately identified as different sources of authority. The differences are legitimate, but cannot be collapsed into one unqualified ``approved'' flag. For a procurement order, a manager's approval of a specific amount is not equivalent to session-wide permission to create orders. A gateway policy's authorization is not the same event as human consent. And a timestamped log of an invocation does not on its own establish whether an external effect occurred. \\

\begin{table}[tbp]
\caption{Different decision meanings across agent harnesses and enterprise platforms.}\label{tab:hgc1}
\centering
\footnotesize
\setlength{\tabcolsep}{4pt}
\renewcommand{\arraystretch}{1.13}
\begin{tabularx}{\linewidth}{@{}N{0.1\linewidth} H H@{}}
\toprule
\textbf{Design} & \textbf{Decision scope} & \textbf{Decision enforcement} \\
\midrule
DeepSeek Harness & One-time approval for a particular tool call; an existing call identifier can link to the presented request. & The harness approval service must return an explicit one-time grant before continuation. \\
OpenAI Codex & An individual command or a category of commands for a session; the prompt also describes requesting a persistent prefix rule. & The approval policy works alongside a sandbox. Model-facing instructions about prefix requests are not proof of enforcement. \\
Anthropic Claude Code & Configurable \texttt{allow}, \texttt{ask}, and \texttt{deny} rules apply to matching tool calls. & Claude Code evaluates permissions before execution. When enabled, the OS-level shell sandbox separately restricts filesystem and network access \cite{ref5, ref11}. \\
Amazon AgentCore & A policy decision for a tool invocation, based on caller and request; it is not human consent. & The gateway blocks denied tool calls in ENFORCE mode. LOG\_ONLY observes decisions without blocking them. \\
Microsoft toolkit ADR 0030 & A proposed approval for an exact action digest, with policy and approval-chain versions and expiry. & Its design calls for checks at execution and one-time consumption where required; an accepted ADR is not proof of deployment. \\
\bottomrule
\end{tabularx}
\end{table}

\noindent\textbf{Policy coverage and failure recovery matter too.} AgentCore's documentation specifies gateway-policy coverage for MCP tool calls, while MCP prompt and resource requests are outside that policy evaluation; this statement does not imply those requests lack other access controls \cite{ref8}. A compatibility claim therefore needs to name which effects are actually mediated. DeepSeek Harness also distinguishes a call not recorded as started from one that started but has no durable result. The latter is reported to the model as having an unknown outcome, with advice against an unsafe retry \cite{ref18}. That warning is valuable, but the host cannot depend solely on a new model following textual advice after handover. \\

\noindent\textbf{What this comparison establishes.} HGC does not replace sandboxes, identity systems, enterprise policy engines, or logging tools. It specifies the boundaries between them: who authorizes an action; the exact action that was authorized; whether dispatch may already have happened; the evidence for an external result; and the state a replacement runtime must inherit. 

\section{The Boundary HGC Governs}\label{the-boundary-hgc-governs}

\subsection{Two roles and one authority}\label{two-roles-and-one-authority}

The \defn{host} is accountable for policy, authorized access, decisions, and the authoritative record of governed work. It can delegate enforcement to trusted services but remains responsible for their behavior. The \defn{runtime} decides how the work gets done. It may use a language model, keep memory, and call local helpers; HGC does not govern how it reasons or what it does internally. When the runtime wants to act on a protected resource, it sends the exact request to the host, and the host or a trusted delegate checks authorization and carries out the request. The runtime must have no direct route that bypasses this check.
Each run acts for a \defn{principal}: an authenticated person or service. A \defn{provider} supplies the runtime, model, or both. A \defn{target} is the protected system affected by an action, such as a procurement service, code repository, or business database. The term host describes the governance role here; it need not be the component called a host in a transport protocol.
A \defn{run} is a defined unit of work with a specified scope of authority. A \defn{grant} gives standing permission to request a bounded operation. A \defn{capability} names an operation that a runtime or service can support, not a grant of authority to use it. A \defn{capability profile} defines the operation's version, inputs, destinations, effects, preconditions, and recovery needs. For example, creating a purchase order is a capability; the employee's procurement limit is a grant; an individual manager's approval is a decision about one proposed order.
HGC governs actions that read protected resources, disclose protected data, or have persistent or external effects; the line these actions cross is the \defn{governed boundary}. Private reasoning need not cross this boundary. Sending a confidential document to a cloud model does cross it, even if the model never invokes a conventional tool. Accepting a provider does not automatically authorize every disclosure to that provider.

\subsection{The host stays while the runtime changes}\label{the-host-stays-while-the-runtime-changes}

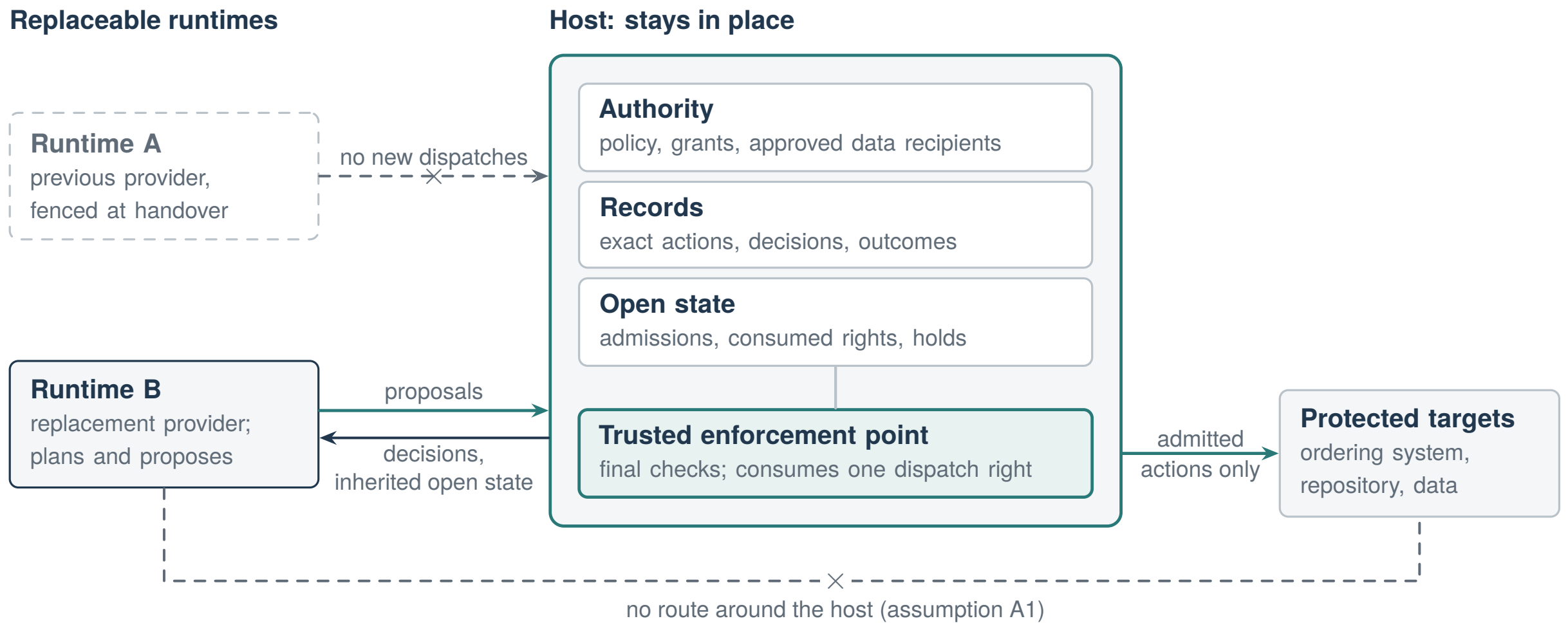
\begin{figure}[tbp]
\centering
{\sffamily\hyphenpenalty=10000\exhyphenpenalty=10000
\begin{tikzpicture}[hgc]
\tikzset{panel/.style={box, fill=white, text width=4.85cm, minimum height=0.86cm}}
\def\hx{8.45}
\node[panel] (auth) at (\hx, 2.10) {\hgcHead{Authority}\\ \hgcNote{policy, grants, approved data recipients}};
\node[panel] (rec)  at (\hx, 1.10) {\hgcHead{Records}\\ \hgcNote{exact actions, decisions, outcomes}};
\node[panel] (open) at (\hx, 0.10) {\hgcHead{Open state}\\ \hgcNote{admissions, consumed rights, holds}};
\node[key, text width=4.85cm, minimum height=0.86cm] (pep) at (\hx,-1.25)
  {\hgcHead{Trusted enforcement point}\\ \hgcNote{final checks; consumes one dispatch right}};
\begin{scope}[on background layer]
  \node[draw=HGCTeal, fill=HGCPanel, rounded corners=4pt, line width=0.9pt,
        inner xsep=8pt, inner ysep=8pt, fit=(auth)(pep)] (host) {};
\end{scope}
\draw[HGCLine, line width=0.8pt] (open.south) -- (pep.north);
\node[ttl, anchor=south west, inner xsep=0pt] (hostttl) at ([yshift=2pt]host.north west) {Host: stays in place};
\node[ghost, text width=2.75cm, minimum height=1.3cm] (A) at (1.55, 1.60)
  {\hgcHead{Runtime A}\\ \hgcNote{previous provider,}\\ \hgcNote{fenced at handover}};
\node[box, draw=HGCNavy, text width=2.75cm, minimum height=1.3cm] (B) at (1.55,-0.95)
  {\hgcHead{Runtime B}\\ \hgcNote{replacement provider;}\\ \hgcNote{plans and proposes}};
\node[ttl, anchor=west, inner xsep=0pt] at (A.west |- hostttl) {Replaceable runtimes};
\node[box, text width=2.45cm, minimum height=1.3cm] (T) at (14.45,-1.25)
  {\hgcHead{Protected targets}\\ \hgcNote{ordering system,}\\ \hgcNote{repository, data}};
\draw[mflow] (A.east) -- node[lbl, above] {no new dispatches} (A.east -| host.west);
\node[text=HGCMuted, font=\hgcMark] at ($(A.east)!0.5!(A.east -| host.west)$) {$\times$};
\draw[tflow] ([yshift=4pt]B.east) -- node[lbl, above] {proposals} ([yshift=4pt]B.east -| host.west);
\draw[flow] ([yshift=-4pt]B.east -| host.west) -- node[lbl, below] {decisions,\\inherited open state} ([yshift=-4pt]B.east);
\draw[tflow] (pep.east -| host.east) -- node[lbl, above] {admitted} node[lbl, below] {actions only} (T.west);
\draw[mflow, -] (B.south) -- ++(0,-0.95) coordinate (b1) -| (T.south);
\node[fill=white, text=HGCMuted, font=\hgcMark, inner sep=1pt] at (b1 -| host.south) {$\times$};
\node[lbl, anchor=north] at ($(b1 -| host.south)+(0,-0.14)$) {no route around the host (assumption A1)};
\end{tikzpicture}}
\caption{Replace the runtime, keep the host. At handover the host fences Runtime A, so it can start no new dispatches, and Runtime B inherits the host's open state. Governed effects reach protected targets only as admitted actions through the trusted enforcement point; no route bypasses the host (assumption A1, Section~\ref{what-the-safety-argument-assumes}). Giving a new provider protected data requires a separate disclosure decision (H2).}
\label{fig:boundary}
\end{figure}

Figure~\ref{fig:boundary} illustrates the central design choice: the runtime on the left can be replaced, while the host in the middle stays. Both runtimes must use the same trusted path for governed effects. The host also retains the old runtime's pending work, decisions, and unresolved outcomes. Handover does not silently expand which providers may receive protected data.

\subsection{What the safety argument assumes}\label{what-the-safety-argument-assumes}

HGC's protection is conditional on three assumptions. They are deployment obligations, not facts that follow from using the HGC name.

\textbf{A1. All claimed effects pass through the host.} Every governed effect covered by a compatibility claim must cross a trusted, host-controlled enforcement point \cite{ref19, ref20}. The host must withhold credentials, network paths, and delegated routes that would let the runtime bypass that point. The claim does not cover effects outside the path it can actually mediate.

\textbf{A2. The host's state is authoritative and protected.} Identities, policy decisions, action identifiers, dispatch state, and history cannot be silently altered or rolled back. Concurrent checks must have a consistent ordering.

\textbf{A3. Execution state survives crashes.} Admission and the one-time consumption of the action's dispatch right are durable before the request can leave the enforcement point. A restart must not recreate a consumed right.

This threat model does not trust the runtime's explanation of what it has done. Retrieved instructions or tool outputs may be malicious; prompt injection can misuse an agent's apparent authority \cite{ref21, ref22, ref23}. Workers can crash, networks can duplicate or delay messages, and remote services can commit effects before a response reaches the host \cite{ref24}.
The important limit is that at-most-once host-side dispatch is not exactly-once effect at the target. A remote system or intermediary may behave differently from the host's single send. Conversely, the host may consume its dispatch right and crash before sending anything, leaving a safe but unresolved action. HGC deliberately gives priority to not repeating an uncertain action over automatic completion of every action.

\section{How the Contract Works}\label{how-the-contract-works}

HGC connects a proposed action to the authority that allowed it and to the evidence of what happened. It does so without requiring a particular database or transport. The host may store the records directly or make them available by stable, resolvable identifiers; the records are not credentials.

\subsection{Five records that preserve meaning}\label{five-records-that-preserve-meaning}

Table 2 gives the five records. Each answers a question an organization must be able to ask after a provider change. A record may refer to another record, but neither a field name nor an adapter can make two different decisions equivalent.

\begin{table}[tbp]
\caption{The five HGC records. They preserve identity, authority, action content, decisions, and execution evidence. Admission and dispatch consumption are states of the execution record, not further record types.}\label{tab:hgc2}
\centering
\footnotesize
\setlength{\tabcolsep}{4pt}
\renewcommand{\arraystretch}{1.13}
\begin{tabularx}{\linewidth}{@{}N{0.19\linewidth} N{0.26\linewidth} H@{}}
\toprule
\textbf{Record} & \textbf{Question} & \textbf{Essential content and owner} \\
\midrule
Runtime binding & Which runtime was accepted, and for what work? & The provider, versions, supported capability profiles, effects and recovery behavior. The runtime declares support; the host accepts the terms. \\
Governed run & Whose authority is being used? & Principal, goal, scope, grants, policy references, approved recipients, runtime binding, and predecessor links. The host controls changes. \\
Action revision & What exact operation was proposed? & Stable identifiers, operation, target, acting identity, executable inputs, destination, preconditions, and effects. The host fixes or validates its contents. \\
Decision & Why could this revision proceed? & Allow, deny, or wait; authenticated decision source, policy version, validity conditions, and expiry. The host issues the authoritative decision. \\
Execution record & What was dispatched, and what is known about its outcome? & Unique action and admission keys, ownership, dispatch state, outcome, evidence, uncertainty holds, and reconciliation. The host owns this history. \\
\bottomrule
\end{tabularx}
\end{table}

For a purchase order, the \defn{action revision} identifies the supplier, amount, delivery address, acting identity, and any other facts material to authorization. If the agent changes the amount after a manager approves, the host creates a new revision and requires a new decision. A link to a mutable attachment must either identify immutable content or be checked again before dispatch.

Each attempt has an \defn{action key}, \(k = (r,a,v)\), consisting of run \(r\), logical action \(a\), and revision \(v\). The host assigns or validates these identifiers and rejects the same key with different executable content. Resending a proposal uses the original key; intentionally repeating an operation requires a separately authorized, linked revision. Keys alone cannot recognize every semantic duplicate. A replacement agent might describe the same business action using a new key, which is why the host also needs the host-enforced holds explained in Section 4.3 \cite{ref2, ref26}.

\subsection{Approval is not dispatch}\label{approval-is-not-dispatch}

\begin{figure}[tbp]
\centering
{\sffamily\hyphenpenalty=10000\exhyphenpenalty=10000
\begin{tikzpicture}[hgc]
\def\mx{4.05}   
\def\rx{12.45}  
\tikzset{stp/.style={box, text width=5.55cm, minimum height=0.95cm},
         outc/.style={box, text width=4.45cm}}
\node[stp] (s1) at (\mx, 0.00) {\hgcHead{1\enspace Resolve the exact action}\\ \hgcNote{target, inputs, acting identity, data recipient}};
\node[stp] (s2) at (\mx,-1.35) {\hgcHead{2\enspace Decide: allow, deny, or wait}\\ \hgcNote{authority checked separately from approval}};
\node[stp] (s3) at (\mx,-2.70) {\hgcHead{3\enspace Admit the action key durably}\\ \hgcNote{one execution attempt; nothing sent yet}};
\node[stp, key] (s4) at (\mx,-4.45) {\hgcHead{4\enspace Consume the dispatch right}\\ \hgcNote{final checks; one durable, irreversible step}};
\node[stp, key] (s5) at (\mx,-5.80) {\hgcHead{5\enspace Send the request once}\\ \hgcNote{from the trusted enforcement point}};
\foreach \a/\b in {s1/s2,s2/s3,s3/s4,s4/s5} \draw[flow] (\a) -- (\b);
\node[lbl, anchor=west] at ($(s2.south)!0.5!(s3.north)+(0.06,0)$) {allow};
\foreach \top/\bot/\name in {s1/s2/Decide, s3/s3/Admit, s4/s5/Send} {
  \draw[HGCLine, line width=0.9pt] ($(\top.north west)+(-0.22,-0.06)$) -- ($(\bot.south west)+(-0.22,0.06)$);
  \node[ttl, rotate=90, anchor=south] at ($($(\top.north west)!0.5!(\bot.south west)$)+(-0.28,0)$) {\name};
}
\node[outc, fill=white, minimum height=1.7cm] (no) at (\rx,-2.03)
  {\hgcHead{No dispatch}\\ \hgcNote{denied, or cancelled or revoked}\\ \hgcNote{before the right is consumed}};
\node[outc, key, minimum height=1.7cm] (unk) at (\rx,-5.12)
  {\hgcHead{Unknown outcome}\\ \hgcNote{the right stays consumed; the host}\\ \hgcNote{holds conflicting work (H6)}};
\node[outc, minimum height=1.25cm] (est) at (\rx,-7.45)
  {\hgcHead{Established outcome}\\ \hgcNote{succeeded, or failed with any}\\ \hgcNote{partial effects recorded}};
\draw[flow] (s2.east) -- node[lbl, above] {deny} (s2.east -| no.west);
\draw[flow] (s3.east) -- node[lbl, below] {cancel or revoke} (s3.east -| no.west);
\draw[mflow] (s4.east) -- node[lbl, above] {crash after this step} (s4.east -| unk.west);
\draw[tflow] (s5.east) -- node[lbl, below] {no conclusive reply} (s5.east -| unk.west);
\draw[flow] (s5.south) |- node[lbl, above, pos=0.75] {conclusive evidence} (est.west);
\draw[flow] (unk.south) -- node[lbl, right] {reconcile from evidence} (est.north);
\coordinate (nr) at ($(s3.south)!0.5!(s4.north)$);
\draw[HGCTeal, dashed, line width=0.8pt] ($(nr -| s1.west)+(-0.5,0)$) -- (nr -| no.east);
\node[fill=white, text=HGCTeal, font=\hgcSmall\bfseries, inner sep=2pt] at (nr -| \rx,0)
  {point of no return: the right cannot be reopened};
\end{tikzpicture}}
\caption{The life of one action key. A denial, cancellation, or revocation before step 4 means nothing is sent; a wait ends in a later allow or deny. Consuming the dispatch right (step 4) is the point of no return: the right is never reopened. If the send or its result is uncertain, the host records an unknown outcome, holds conflicting work, and resolves it only from evidence.}
\label{fig:lifecycle}
\end{figure}
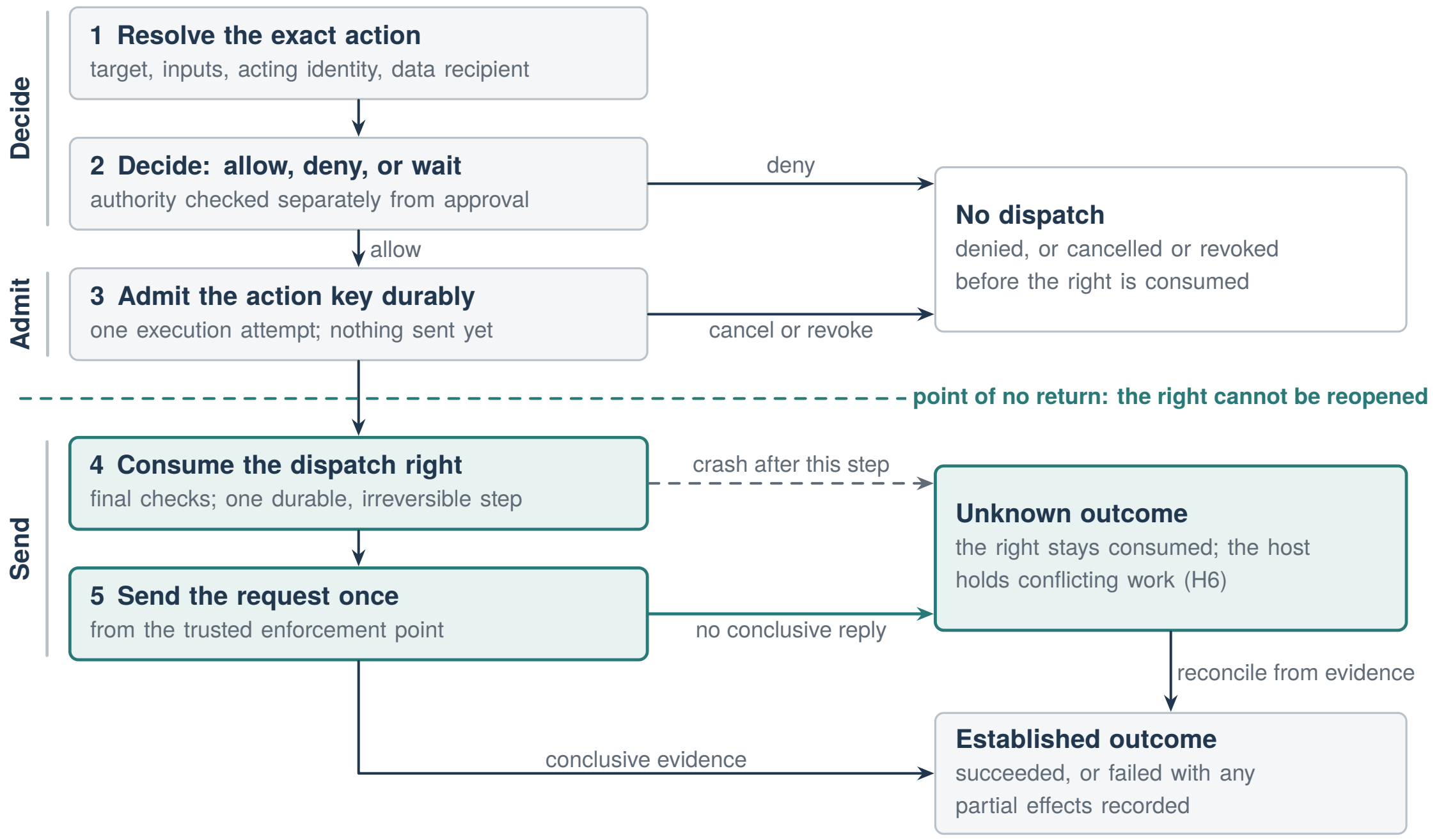

There is a meaningful distinction between deciding that an action is allowed and actually sending it. Figure~\ref{fig:lifecycle} follows one action from proposal to outcome, in three phases: decide, admit, and send.

The runtime proposes an operation; the host resolves its executable content, acting identity, destination, and relevant preconditions. The host then checks standing authority and determines whether an additional approval is required. Its decision is allow, deny, or wait. A wait does not permit dispatch. An approval preview helps a reviewer, but must remain bound to the exact executable revision.

If the action is allowed, the host creates a durable \defn{admission}: a record authorizing one execution attempt. This is not evidence that the request was sent. The admission contains one \defn{dispatch right}, meaning permission for the host to send that specific request at most once. Immediately before sending, the trusted enforcement point rechecks authority, approval, current execution ownership, and unresolved holds. It then consumes the right durably and irreversibly. Here \defn{dispatch} means sending the executable request to the target---not sending the execution record. Only the executor that completes this transition may send the request, and it may do so no more than once. Stale workers must be blocked at the dispatch boundary, not merely prevented from updating the database. Leases can help coordinate workers but cannot substitute for this check \cite{ref27}.

\subsection{An uncertain response is not proof of failure}\label{an-uncertain-response-is-not-proof-of-failure}

After the host sends a request, the target may commit an order or deployment even though no reply reaches the host. Four reporting conditions are useful here: (1)~\defn{not dispatched} means the dispatch right is still unconsumed; (2)~\defn{succeeded} means evidence establishes the required completion; (3)~\defn{failed} means evidence establishes non-completion and records any partial effects; (4)~\defn{unknown} means the host cannot determine whether a potentially sent request took effect. These are not four interchangeable terminal states: ``not dispatched'' concerns execution progress, while the other three concern outcome evidence. An admitted request can still be pending, and a consumed right followed by a crash is treated as unknown even if the request may never have left the host. A timeout or error is not proof of non-execution. The host can resolve uncertainty through an authorized target lookup, target-side idempotency evidence, or attributable operator evidence \cite{ref28}, while retaining the original uncertain observation.

While an outcome is unknown, the host imposes a \defn{hold}: it blocks admission and dispatch of other actions that could duplicate or depend on the uncertain action. This group is its \defn{conflict domain}. The safe default is all governed actions in the run; the host may adopt a narrower domain only when it can validate the capability-specific dependency rules. Authorized human exceptions are possible, but must be recorded and cannot restore a consumed dispatch right. A hold therefore protects more than the original action key: it also constrains a replacement runtime proposing equivalent work under a new key.

\subsection{What handover must carry}\label{what-handover-must-carry}

Table 3 follows an illustrative procurement handover involving two independently developed runtimes. This example also shows why a good contract is not just an audit schema. It connects an earlier approval to an exact action, prevents a changed action from borrowing that approval, and treats an unresolved effect as live state that a replacement must inherit.

\begin{table}[tbp]
\caption{Illustrative provider handover in procurement. The host, not the replacement model's memory or explanation, preserves the pending external effect and controls repetition.}\label{tab:hgc3}
\centering
\footnotesize
\setlength{\tabcolsep}{4pt}
\renewcommand{\arraystretch}{1.13}
\begin{tabularx}{\linewidth}{@{}N{0.037\linewidth} N{0.53\linewidth} H@{}}
\toprule
\textbf{Step} & \textbf{Procurement handover} & \textbf{Host-owned state} \\
\midrule
1 & A purchasing officer starts an authorized run. Runtime A prepares an order within the officer's procurement scope; submission needs a manager's approval. & Runtime binding, run, authorized scope, and policy. \\
2 & The manager approves the order. The runtime changes its delivery address before submission. & A revised action and a new decision; earlier approval no longer applies. \\
3 & The host validates the new approval, consumes the dispatch right, and sends the order. The connection fails before a result is recorded. & Consumed dispatch right, unresolved outcome, and a hold on conflicting orders. \\
4 & The organization switches to Runtime B. Access to any new model recipient is decided separately. Runtime A is fenced from sending new actions. & New runtime agreement; earlier admission, uncertainty, and history remain. \\
5 & Runtime B proposes another order for the same purchase under a new identifier. & The new proposal is held; a fresh identifier does not remove uncertainty. \\
6 & An authorized query finds the submitted order in the purchasing system. The host records the evidence and marks the second proposal as superseded. & Reconciliation evidence and an explicit decision not to send the duplicate. \\
7 & Runtime B continues other authorized work. A genuinely separate order requires its own decision. & Permissions and execution history retain their meaning after handover. \\
\bottomrule
\end{tabularx}
\end{table}

\subsubsection*{How this applies beyond procurement.}

For a coding agent, permission to read source code does not automatically authorize a production deployment. The host checks deployment authority separately, obtains any additional approval required by policy, and treats a changed release target or artifact as a new decision. A lost deployment response requires evidence from the deployment system before conflicting releases proceed. Similarly, a research agent cannot treat a switch of model providers as blanket permission to disclose confidential documents to the new provider. The host authorizes each new recipient and retains the earlier disclosure and decision records across handover. These examples involve different risks, but the same principle: a provider change must not rewrite organizational authority or unfinished work.

\section{Eight Obligations of the Host Governance Contract}\label{eight-obligations-of-the-host-governance-contract}

The previous section explains the mechanism. \textbf{H1--H8 are the eight host obligations} (H stands for host); they state what the trusted host boundary must enforce. Capitalized MUST, MUST NOT, and MAY follow Best Current Practice 14 (BCP 14), which defines standard requirement language \cite{ref29, ref30}. Unless stated otherwise, these obligations apply to the host and any trusted enforcement component. A runtime is expected to cooperate, but host safety must not rely on that cooperation.

\paragraph{H1) Identify the participants and the work they may do.} Before governed work starts, the host MUST authenticate the participants and accept a runtime binding. A governed run MUST record its principal, authorization scope, grants, accepted runtime and provider version, and applicable policy. The runtime MUST declare the HGC and capability-profile versions it supports. If a mandatory requirement is unsupported or unknown, the host MUST NOT activate the affected work. A provider's identity alone MUST NOT grant it access to resources. A later upgrade MUST NOT rewrite which runtime performed earlier work.

\paragraph{H2) Keep capability separate from permission.} The host MUST map each requested operation to an agreed, versioned capability definition, including its resource and disclosure boundaries. It MUST NOT admit an unsupported, ambiguous, or unknown operation. A runtime's claim that it can use a tool MUST NOT be treated as authorization. The runtime MUST NOT use private credentials or undeclared delegation to reach governed targets; the host MUST NOT rely on that promise. It MUST withhold every credential and route that would allow the runtime to bypass host enforcement for a claimed governed effect. The runtime binding MUST disclose effects the host cannot mediate and exclude them from the compatibility claim. Before protected data goes to a new provider, the host MUST authorize that recipient.

\paragraph{H3) Distinguish a person's authority from approval of an action.} For each governed action, the host MUST check the principal's current grants, scope, and policy separately from any action-specific approval. Approval MUST NOT create authority the principal did not have. The record MUST distinguish a human decision from deterministic policy, scoped pre-authorization, and automated review. An automated reviewer MUST NOT satisfy a requirement for human approval. Denial, waiting, missing permission, and an unavailable reviewer MUST NOT count as allow. The runtime MUST NOT manufacture approval evidence; the host MUST authenticate its source and verify the action revision to which it applies.

\paragraph{H4) Approve the exact executable action.} Before admitting an action, the host MUST resolve or validate the operation, target, destination, inputs, acting identity, and relevant preconditions. An allow decision MUST identify the specific revision, decision source, applicable policy and version, and validity conditions. The host MUST reject reuse of one action key with different content. A material change---including a change to referenced content---MUST create a new revision requiring a new decision. Adapters MUST preserve this link, and the enforcement point MUST reject stale or mismatched decisions and verify that the sent content is the authorized content.

\paragraph {H5) Allow at most one host-side send for each action key.} Before dispatch, the host MUST create a durable, unique admission for the action key. The enforcement point MUST atomically check current authority, decision validity, execution ownership, and applicable holds, then durably consume that key's dispatch right before sending the executable request. The enforcement point MUST reject any attempt to send by an executor other than the successful consumer and MUST reject a second send, including one attempted by that consumer. Consumption permits a send but does not prove one occurred. Retries and recovery MUST NOT consume or dispatch the key a second time. Ownership checks MUST fence stale workers at the send boundary, not just prevent stale database writes. The interface binding MUST define ordering against revocation, cancellation, and holds and MUST prevent transparent connector retries from sending the same request again. Consumed keys MUST remain non-reusable. After replay-detail retention expires, old keys MUST still be rejected, for example through a durable anti-replay record or a non-reusable key namespace. A cooperative runtime MAY resubmit the same proposal under its existing key to ask for status; the host MUST return existing state or reject an expired key, never treat that resubmission as permission for another send.

\paragraph {H6) Keep uncertainty visible and prevent unsafe repetition.}

\textbf{(a)} \emph{Record.} If the host cannot establish whether a governed request left the enforcement boundary or had an external effect after the right was consumed, it MUST record the outcome as unknown. The execution record documents the request and observations; it is not the object being dispatched.

\textbf{(b)} \emph{Hold.} Until it is resolved, the host MUST block both admission and dispatch of other actions in the affected conflict domain. After a crash, it MUST recover unresolved consumed admissions and establish the holds before affected work resumes.

\textbf{(c)} \emph{Scope.} The default conflict domain is all governed actions within that run. The host MUST validate any narrower domain against authoritative action data and a defined capability profile, rather than accept only the runtime's description.

\textbf{(d)} \emph{Exceptions.} An authorized person MAY explicitly allow specified work despite uncertainty. The host MUST record the person, authority, scope, and supporting evidence. The exception MUST NOT erase the unknown state or restore a consumed dispatch right. Separately authorized reconciliation reads or transfer steps MAY proceed only if they cannot repeat the uncertain effect or rely on its success. Disclosures remain governed by H2.

\textbf{(e)} \emph{Release.} When evidence resolves an unknown, the host MUST record known effects and reassess each held proposal before release. A deliberate repeat MUST be a newly authorized revision linked to its predecessor.

\textbf{(f)} \emph{Runtime.} The runtime MUST report uncertainty and MUST NOT disguise a retry as unrelated work. The host's own checks must remain effective if it does not cooperate.

\subsubsection*{H7) Preserve an interpretable and protected history.}

The host MUST retain the links among accepted runtime, run, action revision, decision, admission, dispatch state, and outcome. Reports from a runtime, observations from a target, model claims, and reconciliation judgments MUST remain distinguishable. Records MUST NOT reveal reusable credentials. Secret inputs MUST be protected or referenced without losing their link to the executable action. The interface binding MUST define retention, access, redaction, and how missing evidence is represented. An upgrade MUST NOT reinterpret older decisions, restore consumed dispatch rights, or discard unresolved admissions. Missing evidence MUST NOT be treated as proof that an action had no effect.

\paragraph{H8) Make replacement, cancellation, and revocation explicit.} The host MUST distinguish stopping new runs, revoking current authority, pausing or cancelling work, and replacing a runtime. Each operation MUST define its effect on pending, admitted, and already in-flight actions and preserve earlier history. A replacement MUST establish a compatible runtime binding and use an explicit handover or a newly authorized run. Outstanding admissions, consumed keys, and unknown results MUST remain. Continuing known unresolved work in a successor run MUST retain the applicable hold unless a person authorizes an exception under H6(d). Before the new runtime can begin governed dispatch, the host MUST fence the former runtime's ability to start more dispatches; previously sent requests remain tracked. Cancellation MUST NOT be portrayed as undoing an external effect without evidence of a compensating operation \cite{ref31}.

\paragraph {Policy remains the organization's choice.} HGC standardizes the meaning and handling of a decision, not whether the organization should allow a given order, deployment, or document disclosure. A host may require human approval for each external write or use bounded deterministic authorization. Both are consistent with HGC if decision source, action identity, enforcement, and history remain explicit.

\section{What HGC Can Guarantee When a Provider Changes}\label{what-hgc-can-guarantee-when-a-provider-changes}

The value of HGC is not that two agents generate the same plans. It is that a host can replace the runtime without losing the rules under which actions were permitted or the meaning of unfinished work. This section separates the safety claim, the replacement criterion, and the evidence still needed to test them.

\subsection{Three conditional safety properties}\label{three-conditional-safety-properties}

The following arguments depend on assumptions A1--A3 in Section 3.3 and on correct enforcement of H1--H8. They are proof sketches, not machine-checked proofs.

\textbf{Property 1} (A proposal does not grant authority). A governed host-side dispatch has an admitted action revision and a valid host decision. The runtime cannot create this authority merely by proposing an operation.

\emph{Proof sketch.} A1 requires all claimed effects to pass through the trusted enforcement point. H5 requires it to check authority and the decision before consuming the dispatch right. H3 separates standing permission from action-specific approval; H4 binds the decision to the resolved action. A runtime proposal alone passes none of these checks. \(\square\)

\textbf{Property 2} (Approval does not cover changed content). A decision for one revision does not authorize another whose authorization-relevant content differs.

\emph{Proof sketch.} H4 rejects an action key reused with different content and requires a new decision for a material change. H5 sends only the revision linked to the admitted decision. A mutable reference or adapter therefore cannot quietly turn an approved order for one amount into authority for another, provided the host validates the actual executable content. \(\square\)

\textbf{Property 3} (Recovery does not restore used dispatch rights). Each action key permits at most one host-side dispatch. Unknown outcomes and consumed rights survive recovery and provider handover. Once a hold has been established, conflicting work cannot be newly admitted or sent without a scoped, recorded human exception until reconciliation.

\emph{Proof sketch.} A2--A3 and H5 make dispatch-right consumption durable, ordered, and non-reusable; stale workers cannot acquire a new right for the same key. H6 requires the host to reconstruct holds before restarting affected work, including proposals with new keys in the same validated conflict domain. H8 preserves those conditions across handover or known successor work. The argument does not recall requests already sent and depends on valid conflict-domain definitions. \(\square\)

These properties do not guarantee exactly-once effects at a remote target \cite{ref24,ref28}. A target might process one request more than once, a network request might already be in flight when authority is revoked, and two unrelated runs might independently propose equivalent orders. HGC also cannot control what an approved provider later does with data it was legitimately permitted to receive. The guarantees cover a declared and properly mediated boundary, not an entire organization merely because its agents use the HGC terminology.

\subsection{Six conditions for governance-preserving replacement}\label{six-conditions-for-governance-preserving-replacement}

For a declared set of capabilities, a replacement runtime preserves governance only if the host accepts the new runtime and these six conditions continue to hold:

\begin{enumerate}
\def\labelenumi{\arabic{enumi}.}
\item
  \textbf{Operations still mean the same thing.} Capability names and versions retain the meanings on which organizational policy depends.
\item
  \textbf{Authority does not expand silently.} New representations cannot widen the principal's grants, resource scope, or approved data recipients.
\item
  \textbf{Execution identity survives.} Approved revisions, action keys, and consumed dispatch rights remain interpretable and non-reusable.
\item
  \textbf{Decisions retain their basis.} The host can still identify who or what decided, under which policy, and for how long the decision is valid.
\item
  \textbf{Outcomes retain their uncertainty.} A failure, partial effect, or unknown result does not become a different status merely because another runtime reads it.
\item
  \textbf{Handover preserves unfinished work.} Pending admissions, holds, history, and the exclusion of the old executor survive the change.
\end{enumerate}

A replacement may support fewer capabilities and may require a different action format; the host must document and validate the mapping. It may reject a provider on data-location, confidentiality, or other policy grounds. A new provider's access to sensitive documents is a fresh disclosure decision, even when its tool API is compatible.

A runtime that simply refuses all requests might satisfy many safety restrictions but would not provide useful interoperability. Any future compatibility claim should therefore include positive traces in which supported, authorized work completes, as well as tests showing unsafe work is blocked.

\subsection{How the requirements could be tested}\label{how-the-requirements-could-be-tested}

Table 4 proposes a conformance suite. Each case must observe both the trusted dispatch boundary and the resulting host state; a runtime's assertion that an action was denied is insufficient if a bypassed tool actually changed the target.

\begin{table}[tbp]
\caption{Proposed HGC conformance cases. Required observations include the host's durable records and an independent view of dispatch or target effects, not only model-generated reports.}\label{tab:hgc4}
\centering
\footnotesize
\setlength{\tabcolsep}{4pt}
\renewcommand{\arraystretch}{1.13}
\begin{tabularx}{\linewidth}{@{}N{0.045\linewidth} N{0.075\linewidth} H H@{}}
\toprule
\textbf{Test} & \textbf{Rule} & \textbf{Stimulus} & \textbf{Required observation} \\
\midrule
C1 & H1 & Offer an unsupported mandatory version. & The affected operation stays inactive; protected data is not sent over that path. \\
C2 & H2 & Advertise an ungranted tool or new data recipient. & The host blocks the operation or makes a distinct, valid authorization decision. \\
C3 & H3 & Approve after revocation; offer automated review where a person is required. & Neither causes dispatch. \\
C4 & H4 & Change approved inputs, destination, identity, or an attachment. & The old decision does not authorize the changed action. \\
C5 & H5 & Race two sends and resume a stale worker. & One consumption and at most one host-side dispatch per key. \\
C6 & H5 & Crash before consumption, between consumption and send, and after send. & Only a still-unconsumed key can proceed normally. Consumed uncertain keys are not resent. \\
C7 & H5 & Replay an expired key or trigger connector auto-retry. & The old key is rejected; no second host dispatch appears. \\
C8 & H6 & Lose a result, replace the runtime, propose equivalent work under a new key. & The unknown outcome and conflict hold persist. \\
C9 & H6 & Produce partial effects, then reconcile while another proposal is held. & Partial effects are recorded; held work is reassessed, not released automatically. \\
C10 & H6--H8 & Continue an unknown operation through handover or a successor run. & Holds and consumed rights remain; new disclosures need authorization. \\
C11 & H7 & Reconcile with operator evidence, then redact details under retention rules. & Attribution remains; missing details are not treated as proof of non-execution. \\
C12 & H8 & Race revocation, cancellation, and handover against dispatch. & Ordering is respected; old ownership is fenced; in-flight effects remain recorded. \\
C13 & H1--H8 & Complete an allowed action, query the result again, and hand over. & The result can be retrieved without a duplicate effect; the replacement can continue supported work. \\
\bottomrule
\end{tabularx}
\end{table}

The evaluation should retain ordered event traces, target observations, and final host state. Tests need both failure injection and ordinary successful operations to measure safety and useful continuity.

\section{Relationship to Existing Work}\label{relationship-to-existing-work}

HGC draws on established ideas in access control, secure execution, and distributed systems. Its proposed contribution is not the invention of one-time approval, audit records, or crash recovery. It is an agreement about which of these properties a host must preserve when an independently developed agent runtime changes.

\paragraph{Connecting agents and tools.} MCP, A2A, and the Agent Client Protocol standardize different forms of connection: tools and resources, agent-to-agent communication, and coding-agent/editor integration \cite{ref1,ref2,ref3,ref50}. Agent Plugins addresses packaging \cite{ref32}, with its future-considerations document discussing additional concerns such as provenance, permissions, and audit \cite{ref33}. HGC can be carried over such interfaces, but compatibility at the messaging layer does not itself establish whether an old approval or unresolved effect keeps its meaning.

\paragraph {Approving and enforcing agent actions.} The Agent Action Decision Protocol (AADP) specifies per-action decisions, approval handling, and conservative behavior when an external effect is uncertain \cite{ref34}. Microsoft's Agent Governance Toolkit ADR 0030 proposes action-bound approval with execution-time validation and one-time use \cite{ref35}. The Agent Execution Protocol describes an agent loop governed by an enforcement component \cite{ref36}; the Verifiable Agent Protocol uses intent-linked admission and audit \cite{ref37}; and Fernandez's Agent Control Protocol addresses admission and execution records, including cross-organization exchanges \cite{ref38}.
AADP is the closest conceptual neighbor. It defines how an action gets an authoritative decision and how its result is reported. HGC instead focuses on the surrounding host--runtime relationship: accepted runtime identity and scope, persistent evidence across a runtime change, the restriction on other conflicting actions while one outcome is unknown, and a six-condition replacement test. They could be used together: an interface binding might use AADP for the decision and dispatch exchange while HGC defines handover and continuity around it.
CapLease, a proposed durable authorization-consumption mechanism, addresses repeat authorization under fresh token identifiers; ACRFence addresses repeated irreversible actions after checkpoint restoration \cite{ref25,ref26}. They reinforce why an action key alone cannot prevent every business-level duplicate. HGC's hold on conflicting work extends the concern to a runtime replacement, while leaving cross-run semantic equivalence to organizational policy.

\paragraph{Policy, identity, and execution foundations.} The distinction between making and enforcing an access decision is reflected in the eXtensible Access Control Markup Language (XACML), an access-control policy standard, and the principle of complete mediation \cite{ref19,ref20}. OAuth Rich Authorization Requests can express detailed authorization, while Token Exchange can represent delegation and acting identities \cite{ref39,ref40}. AgentSpec, Progent, CaMeL, and AIRGuard explore runtime restrictions and the separation of trusted authority from untrusted content \cite{ref23,ref41,ref42,ref43}. These mechanisms may help build a compliant host. HGC does not replace them or claim that specifying records stops prompt injection.
RPC failure semantics explain the uncertainty after a missing response \cite{ref24}. Leases can coordinate execution ownership but do not by themselves stop stale dispatch \cite{ref27}. Idempotent operations can support safe reconciliation where the target provides suitable guarantees \cite{ref28}; sagas show why cancellation is not the same as undoing a completed effect \cite{ref31}. Contract theory and governed interaction motivate describing shared behavioral expectations without requiring common internal implementations \cite{ref44,ref45,ref46,ref47}. Agent Contracts examines resource-bounded agent execution \cite{ref48}, and work on agent visibility motivates monitoring and traceability \cite{ref49}. HGC brings these mechanisms together to preserve organizational control across runtime handover.

\section{From a Design to Working Interoperability}\label{from-a-design-to-working-interoperability}

HGC is intended to be independent of the protocol that carries an agent request. To implement it, three layers are needed.

\subsubsection*{The core contract.}

The five records and H1--H8 define the shared meaning of authority, action decisions, host-side dispatch, recovery, and handover. They say what must stay true, without requiring a particular storage engine or model.

\paragraph{An interface binding.} This is the concrete mapping onto MCP, A2A, the Agent Client Protocol, AADP, or an ordinary service API. It must specify authenticated identities, record identifiers, serialization, executable-action comparison, dispatch and revocation ordering, replay retention, cancellation, evidence retrieval, and failure reports. Calling a protocol endpoint ``HGC-compatible'' is not enough if the mapping changes the behavior.

\paragraph {Capability profiles.} These describe individual domain operations. A procurement profile may define which order fields affect approval, how target-side order identifiers can be verified, and what other purchase actions conflict with an unresolved order. A deployment profile would identify the affected environment and the evidence needed to establish its result. Profiles also specify data destinations, preconditions, effects, and narrower conflict domains where safe. If the host cannot understand or enforce a requirement that an operation depends on, it must block that operation rather than proceed without the check.

Observational rollout can help assess a policy before enforcement, but log-only mode must not be presented as a security boundary \cite{ref16}. Execution traces may use existing telemetry formats \cite{ref6,ref17} as long as they preserve the links to authoritative decisions and evidence rather than replacing them with unverified runtime descriptions.

\paragraph{Versions and future evaluation.} Participants must declare the versions of the HGC contract, interface binding, and capability profiles. Implementations can ignore optional metadata, but not unknown mandatory restrictions. A new version cannot silently reinterpret an old decision or restore a used dispatch right.

Future evaluation should implement a public interface binding and executable conformance tests, then examine at least two independently developed hosts and two independently developed runtimes. Trials should cover authorized completion and injected failures, and measure dispatch latency, durable-write overhead, recovery time, and time blocked by uncertainty. Independence matters: two adapters built from one shared implementation do not establish interoperability between different implementations.

\section{Limitations and Open Questions}\label{limitations-and-open-questions}

\paragraph {Mediation can be incomplete.} A local coding agent with broad shell or network access may be able to reach systems without the host. A sandbox configured by that agent does not automatically create independent host control \cite{ref5,ref6}. HGC's safety properties apply only to effects for which the deployment can actually enforce the trusted boundary. They also do not control later use of data already disclosed to an approved provider.

\paragraph{Conservative recovery can block useful work.} A run-wide hold may delay unrelated tasks. Narrowing it requires accurate dependency definitions and domain evidence. A consumed dispatch right cannot be reused even if a crash happened before the request was sent, so an action may remain incomplete. A hold cannot recall in-flight requests, and two unrelated runs might still create equivalent effects. HGC favors explicit uncertainty over an unsafe claim of completion.

\paragraph{Delegation and human judgment need more work.} The current design specifies one host--runtime relationship. Multi-agent delegation must either use the same trusted enforcement path or establish a distinct governed relationship; attenuating grants and coordinating holds across hosts need further specification. A precise approval record cannot prevent rushed human approval or a misleading preview, and automated review can be wrong. The contract does not guarantee eventual progress: practical deployments need escalation, expiry, and authorized recovery procedures that do not turn silence into permission.

\section{Conclusion}\label{conclusion}

Replacing an AI agent should not silently replace an organization's rules. A compatible tool interface does not tell us what an earlier approval covered, whether a request already took effect, or what a new provider is allowed to receive.
The Host Governance Contract proposes a way to keep those answers with the host. Five records and eight obligations link authority to specific actions, controlled dispatch, evidence, and unfinished work. A six-condition replacement criterion describes when those meanings can survive a runtime change. The claims depend on complete, trusted mediation and crash-safe state; they do not promise exactly-once effects at external services.
Its intended value is clear: organizations should be able to choose or replace the agent that performs a task without surrendering control of what it is allowed to do or losing the meaning of what already happened.

\end{document}